%% file: 0_MAIN.tex
\pdfoutput=1

\documentclass[11pt]{article}

\usepackage[final]{acl}

\usepackage{times}
\usepackage{latexsym}

\usepackage[T1]{fontenc}

\usepackage[utf8]{inputenc}

\usepackage{microtype}

\usepackage{inconsolata}

\usepackage{graphicx}
\usepackage{lipsum}  
\usepackage{amsmath}
\usepackage{array}
\usepackage[skins]{tcolorbox}
\usepackage{multirow}
\usepackage{arydshln}
\usepackage{csquotes}
\usepackage{dashrule}
\usepackage{tikz}
\usepackage{booktabs}
\usepackage{makecell}

\usetikzlibrary{shapes.geometric, arrows.meta, positioning, intersections, decorations.pathreplacing, calc, matrix}

\usepackage{tabularray}

\definecolor{DeezerPurple}{HTML}{A238FF}
\definecolor{DeezerOrange}{HTML}{FF673D}
\definecolor{DeezerTeal}{HTML}{96F9F3}
\definecolor{DeezerAcid}{HTML}{FCFF60}
\definecolor{DeezerRed}{HTML}{FF0000}
\definecolor{DeezerTurbo}{HTML}{FFED00}
\definecolor{DeezerRose}{HTML}{FF0092}
\definecolor{DeezerLime}{HTML}{C2FF00}
\definecolor{DeezerRobinsEggBlue}{HTML}{00C7F2}
\definecolor{DeezerFrenchPass}{HTML}{C1F1FC}
\definecolor{DeezerTidal}{HTML}{EBFFAC}
\definecolor{DeezerCottonCandy}{HTML}{FFC2E5}
\definecolor{DeezerCornflowerLilac}{HTML}{FFAAAA}
\definecolor{DeezerWhite}{HTML}{FFFFFF}
\definecolor{DeezerSilver}{HTML}{BFBFBF}
\definecolor{DeezerGray}{HTML}{808080}
\definecolor{DeezerMineShaft}{HTML}{404040}
\definecolor{DeezerPalePurple}{HTML}{EAD1FF}
\definecolor{DeezerMauve}{HTML}{D09AFF}
\definecolor{DeezerElectricViolet}{HTML}{A238FF}
\definecolor{DeezerWindsor}{HTML}{6E14BD}
\definecolor{DeezerPalePink}{HTML}{FDDDFF}
\definecolor{DeezerWhitePointer}{HTML}{FBBBFF}
\definecolor{DeezerHeliotrope}{HTML}{F673FF}
\definecolor{DeezerSeance}{HTML}{C01FC3}
\definecolor{DeezerPaleRed}{HTML}{FFC3C3}
\definecolor{DeezerCoralRed}{HTML}{FF3D3D}
\definecolor{DeezerThunderbird}{HTML}{D71F14}
\definecolor{DeezerPaleOrange}{HTML}{FFDAC6}
\definecolor{DeezerFlesh}{HTML}{FFBB95}
\definecolor{DeezerOutrageousOrange}{HTML}{FF673D}
\definecolor{DeezerPunch}{HTML}{DB452C}
\definecolor{DeezerPaleAcid}{HTML}{FBFFB7}
\definecolor{DeezerPaleCanary}{HTML}{FDFF95}
\definecolor{DeezerLaserLemon}{HTML}{FCFF60}
\definecolor{DeezerCitron}{HTML}{90931E}
\definecolor{DeezerPaleTeal}{HTML}{ECFFFE}
\definecolor{DeezerPaleCanary1}{HTML}{C7FBF8}
\definecolor{DeezerFoam}{HTML}{96F9F3}
\definecolor{DeezerCalypso}{HTML}{2F7C90}
\definecolor{DeezerPaleBlue}{HTML}{E4E7FF}
\definecolor{DeezerPeriwinkle}{HTML}{AAB2FF}
\definecolor{DeezerBlue}{HTML}{3448FC}
\definecolor{DeezerCalypso1}{HTML}{2836B5}

\title{Double Entendre:\\Robust Audio-Based AI-Generated Lyrics Detection via Multi-View Fusion}

\author{
  \begin{tabular}{cccc}
    Markus Frohmann~$^{1,2}$ & Gabriel Meseguer-Brocal~$^{1}$ & Markus Schedl~$^{2,3}$ & Elena V. Epure~$^{1}$ \\
  \end{tabular}
  \\
  $^1$ Deezer Research, Paris, France \\
  $^2$ Johannes Kepler University Linz, Austria \quad
  $^3$ Linz Institute of Technology, AI Lab, Austria \\
  \texttt{research@deezer.com}
}

\begin{document}
\maketitle
\begin{abstract}
    The rapid advancement of AI-based music generation tools is revolutionizing the music industry but also posing challenges to artists, copyright holders, and providers alike.
    This necessitates reliable methods for detecting such AI-generated content.
    However, existing detectors, relying on either audio or lyrics, face key practical limitations:
    audio-based detectors fail to generalize to new or unseen generators and are vulnerable to audio perturbations; lyrics-based methods require cleanly formatted and accurate lyrics, unavailable in practice.
    To overcome these limitations,
    we propose a novel, practically grounded approach: a multimodal, modular late-fusion pipeline that combines automatically transcribed sung lyrics and speech features capturing lyrics-related information within the audio.
    By relying on lyrical aspects directly from audio, our method enhances robustness, mitigates susceptibility to low-level artifacts, and enables practical applicability.
    Experiments show that our method, \textsc{DE-detect}, outperforms existing lyrics-based detectors while also being
    more robust to audio perturbations. Thus, it offers an effective, robust solution for detecting AI-generated music in real-world scenarios.\footnote{Our code 
    is available at \url{https://github.com/deezer/robust-AI-lyrics-detection}.}
\end{abstract}

\section{Introduction and Background}
\label{sec:introduction}
\input{1-introduction}

\section{Method}
\label{sec:method}
\input{2-method}

\section{Experimental Setup}
\label{sec:setup-experiments}
\input{3-setup}

\section{Experiments}
\label{sec:results}

\input{4-results}

\section{Conclusion}
\label{sec:conclusion}
\input{9-conclusion}

\section*{Limitations}
While our multi-view method demonstrates promising results in AI-generated lyrics detection, we acknowledge several limitations that warrant further investigation in future work.

First, our model's training and evaluation are primarily based on the dataset by \citet{labrak2024detectingsyntheticlyricsfewshot}. This introduces potential biases related to the dataset's distribution, despite its inspiration from multiple language and music genres pairs.
We thus encourage future work to introduce and explore larger, more diverse datasets encompassing a wider range of music styles and languages.

Furthermore, relying on Suno v3.5 for generating AI-generated audio for training introduces a potential bias toward this specific tool's artifacts and stylistic characteristics.
Although we evaluated our method on Udio as an out-of-domain generator, our core training remains Suno-centric.
Once other music-generation tools that support lyrics conditioning are available, future research should investigate training and evaluating audio from a more diverse set of AI music-generation tools to reduce tool-specific biases.

We also acknowledge that our robustness evaluation does not cover every potential attack; for instance, attacks that combine two or more audio perturbations (e.g., changing pitch and time stretching). We leave this to future work.

\section*{Ethical Considerations}
While intended for positive applications like copyright protection and transparency, revealing vulnerabilities in detection systems carries a dual-use risk.  Malicious actors could exploit these weaknesses to create AI music designed to evade detection, potentially enabling further copyright infringement and music streaming platform manipulation.
This risk is compounded by the potential for bias in our approach since our model may inherit biases from the training data, leading to unfair or inaccurate detection
\citep{barocas2017problem}.
This could result in unjust content takedown or censorship, disproportionately impacting certain artists \citep{doi:10.1177/27523543241269047}.

Therefore, we advocate for the responsible development and deployment of AIGM detection technologies, emphasizing transparency, fairness, and human-in-the-loop approaches
to maximize benefits while mitigating possible harms to artists, creators, and the broader music ecosystem.

\section{Acknowledgements}
This research was funded in whole or in part by the Austrian Science Fund (FWF):  \url{https://doi.org/10.55776/COE12}, \url{https://doi.org/10.55776/DFH23}, \url{https://doi.org/10.55776/P36413}.
The authors would like to thank Aurelien Herault, Manuel Moussallam, Romain Hennequin, Yanis Labrak, and Gaspard Michel for their invaluable feedback on this work.

\bibliography{ACL2025,neurips_2024}

\clearpage
\appendix
\input{999-appendix}
\label{sec:appendix}

\end{document}

%% file: 1-introduction.tex
\input{figures/pipeline}

The advent of AI-generated music (AIGM) has recently been transformative for the music industry, mainly driven by music generation tools such as Suno\footnote{\href{https://www.suno.com/}{www.suno.com}} or Udio\footnote{\href{https://www.udio.com/}{www.udio.com}}.
While such tools can enhance creativity by aiding in composition and arrangement~\citep{Li2024FromAD, Parada-Cabaleiro2024SciRep_2024}, they also raise concerns regarding copyright, artistic value, and the potential for AI-created works to overshadow human musicians~\citep{Afchar2024DetectingMD, aimusic, doi:10.1177/27523543241269047}.
The divergent responses from music streaming services, with some ceasing to recommend AI-flagged songs\footnote{\href{https://www.billboard.com/pro/deezer-ai-detection-tool-10-percent-music-tracks-ai-generated/}{www.billboard.com/pro/deezer-ai-detection-tool-10-percent-music-tracks-ai-generated}} and others embracing them\footnote{\href{https://www.bigtechnology.com/p/spotifys-plans-for-ai-generated-music}{www.bigtechnology.com/p/spotifys-plans-for-ai-generated-music}}, underscore the increasingly critical need for robust and reliable AIGM detection methods.

Existing work on AIGM detection has mostly focused on AI-generated audio, whether with or without vocals~\citep{Afchar2024DetectingMD,Cooke2024AsGA,Rahman2024SONICSSO}.
While such detectors have been shown to achieve high (>99\%) accuracy within their training domain, they fail to generalize to unseen AIGM models and are highly vulnerable to audio attacks such as adding noise or changing pitch~\citep{Afchar2024DetectingMD}.
This highly limits their usability in practice.

Beyond audio, for songs with vocals, lyrics (represented as text) are an essential medium of conveying a song's content \citep{Li2024FromAD}.
In most AIGM, lyrics are also generated by AI; thus, determining lyrics authorship (human or AI) could be a proxy for flagging a track as AI-generated.
To detect AI-generated \textit{lyrics}, \citet{labrak2024detectingsyntheticlyricsfewshot} introduce a dataset of synthetic lyrics generated using several LLMs, based on prompts informed by lyric examples from diverse language–music genre pairs.
They evaluate various text-based detectors, showing promising results.
However, their methods rely on clean, perfectly formatted lyrics; but in practice, only audio is available, making this requirement impractical.\footnote{In practice, lyrics metadata is often unavailable for newly ingested music in industrial settings.}

\vspace{1mm}
\noindent\textbf{Contributions.}
To overcome these limitations, we propose a novel multi-view pipeline for detecting AI-generated lyrics that is both robust and practically applicable, relying solely on audio as input.
As Figure~\ref{fig:pipeline} shows, it robustly leverages this input as two different modalities: (i) automatically transcribed lyrics to eliminate reliance on perfectly formatted lyrics, and (ii) using speech models to capture lyrics-related information present only in singing voice.
Experiments show that our method exhibits improved, more robust performance than unimodal ones, especially out-of-domain.
This results in a practical solution for robust AI-generated lyrics detection, paving the way for greater transparency in the rapidly evolving AIGM landscape.

%% file: figures/pipeline.tex
\begin{figure*}[ht]
\centering
\begin{tikzpicture}[
    block/.style = {draw, rectangle, minimum width=2.7cm, minimum height=0.6cm, align=center, font=\sffamily\tiny, text=black, fill=gray!20, rounded corners=3pt, inner sep=2pt},
    decision/.style = {draw, rectangle, minimum width=1.8cm, minimum height=0.8cm, align=center, font=\sffamily\small, text=black, fill=gray!20, rounded corners=3pt, inner sep=6pt},
    arrow/.style = {-{Latex[length=2.5mm, width=3mm]}, line width=0.7pt},
    feature/.style = {draw, minimum width=0.6cm, minimum height=1.1cm, fill=DeezerTeal, align=center, font=\sffamily\tiny, rounded corners=3pt, inner sep=2pt},
    shortfeature/.style = {draw, minimum width=0.6cm, minimum height=0.95cm, fill=DeezerTeal, align=center, font=\sffamily\tiny, rounded corners=3pt, inner sep=2pt},
    icon/.style = {minimum size=1cm}, %
]

\node (song) at (0,0) {\includegraphics[width=2cm]{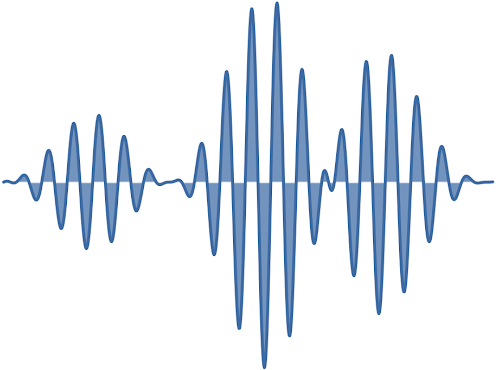}};
\node[above=0.1cm of song, font=\sffamily\bfseries\normalsize] (song-label) {Song};

\draw[arrow] (song.east) -- ++(1.8,-0.9) node[midway, below=0.1cm, font=\sffamily\small, align=center, xshift=-0.4cm] (speech-emb-label) {Speech\\Model} coordinate (speech-emb-arrow);
\node[below=-0.2cm of speech-emb-label] (xeus) {\includegraphics[width=0.75cm]{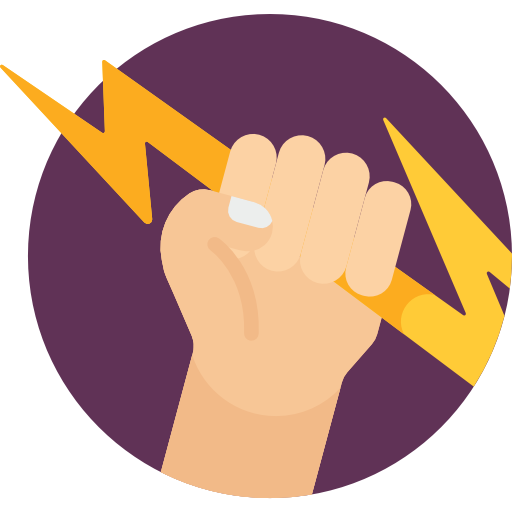}};

\node[block,  inner sep=0pt, right=0.1cm of speech-emb-arrow] (speech-vectors) {
    \begin{tabular}{@{} r @{\hskip 0.5em} r @{\hskip 0.5em} r @{}}
        $\cdot\cdot\cdot$ & $\cdot\cdot\cdot$ & $\cdot\cdot\cdot$  \\
        0.22 & 0.15 & -0.30 \\
        -0.10 & 0.45 & 0.20 \\
        $\cdot\cdot\cdot$ & $\cdot\cdot\cdot$ & $\cdot\cdot\cdot$ \\
    \end{tabular}
};
\node[below=0.2cm of speech-vectors, font=\sffamily\bfseries\normalsize] (speech-vectors-label) {Speech Vectors};

\draw[arrow] (speech-vectors.east) -- ++(1.5,0) node[midway, below=0.1cm, font=\sffamily\small, align=center] (mean-pooling-label) {Mean\\ Pooling} coordinate (mean-pooling-point);

\node[feature, right=0.1cm of mean-pooling-point] (features-top) {
    $\cdot\cdot\cdot $\\
    0.02\\
    0.18\\
    $\cdot\cdot\cdot$
};
\node[below=0.15cm of features-top, font=\sffamily\bfseries\normalsize, align=center, xshift=1cm] (features-label-top) {Speech Features};

\draw[arrow] (song.east) -- ++(1.8,0.9) node[midway, above=0.15cm, font=\sffamily\small, align=center, xshift=-0.4cm] (transcriber-label) {Transcriber} coordinate (transcriber-arrow);
\node[above=-0.25cm of transcriber-label] (transcriber) {\includegraphics[width=1.0cm]{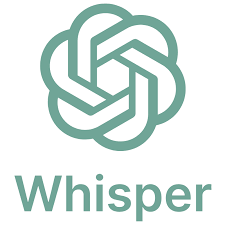}};
\node[midway, right=1.7cm of song] (ice1) {\includegraphics[height=0.4cm]{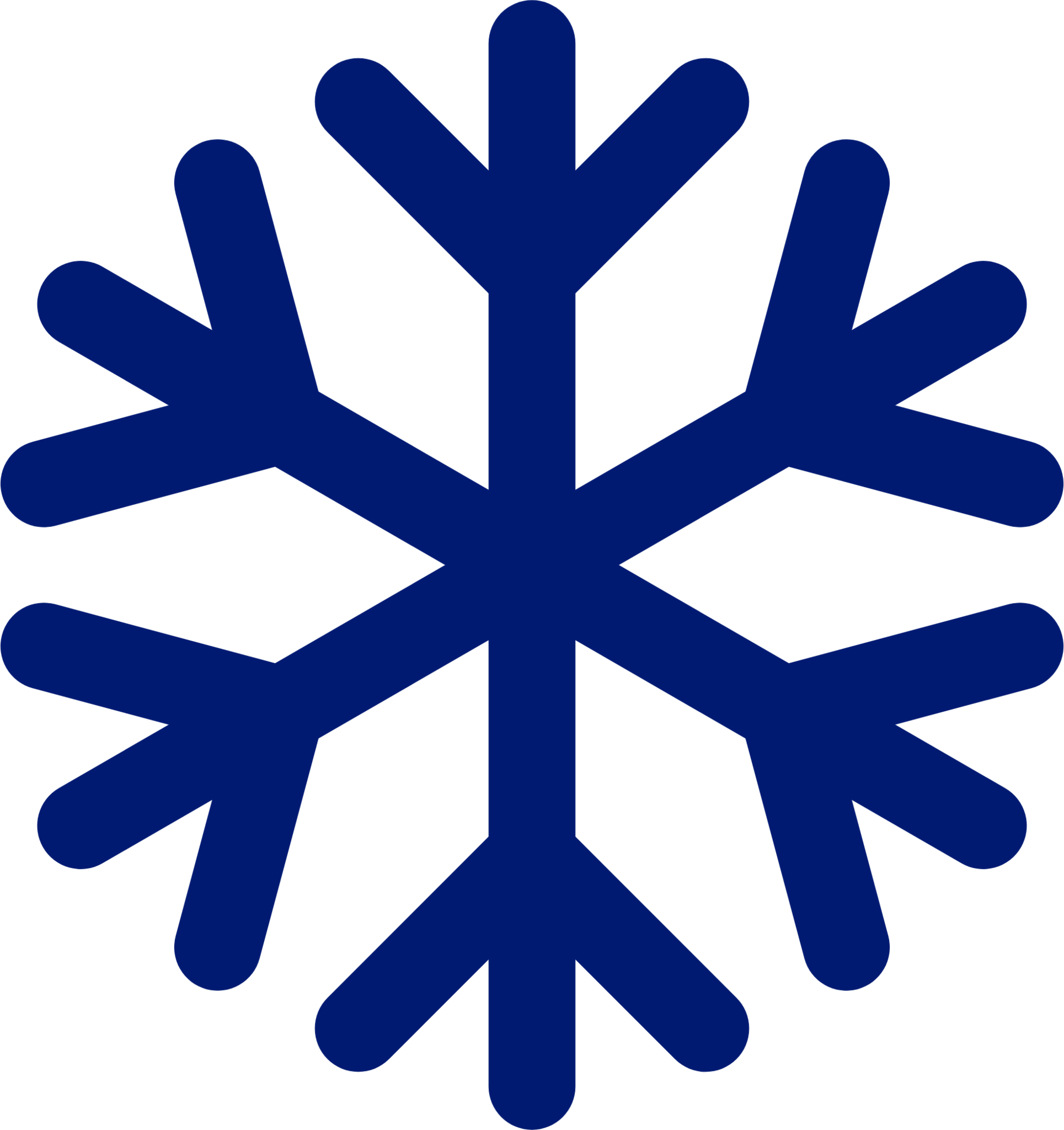}};

\node[block, right=0.1cm of transcriber-arrow, yshift=-0.2cm] (transcription) {"Forever trusting\\ who we were and \\ nothing else shatters"\\...}; %
\node[above=0.2cm of transcription, font=\sffamily\bfseries\normalsize] (transcription-label) {Transcript};

\draw[arrow] (transcription.east) -- ++(1.5,0) node[midway, above=0.1cm, font=\sffamily\small, align=center] (feature-label) {Feature\\ Extraction} coordinate (feature-point);
\node[midway, right=6.15cm of song] (ice2) {\includegraphics[height=0.4cm]{figures/images/ice.png}};

\node[feature, right=0.1cm of feature-point] (features) {
    $\cdot\cdot\cdot$ \\
    0.12\\
    0.03\\
    $\cdot\cdot\cdot$
};
\node[above=0.165cm of features, font=\sffamily\bfseries\normalsize, align=center, xshift=1cm] (features-label) {Text Features};

\node[shortfeature, right=1.2cm of features-top] (inter-top) {
  \begin{tabular}{@{} c @{}}
    $\cdot\cdot\cdot$ \\
    0.45 \\
    $\cdot\cdot\cdot$
  \end{tabular}
};

\node[shortfeature, right=1.2cm of features] (inter-bottom) {
  \begin{tabular}{@{} c @{}}
    $\cdot\cdot\cdot$ \\
    0.05 \\
    $\cdot\cdot\cdot$
  \end{tabular}
};

\draw[arrow] (features-top.east) -- (inter-top.west) node[midway, below=0.12cm, font=\sffamily\small\bfseries, align=center] (speech-mlp) {Linear};
\draw[arrow] (features.east) -- (inter-bottom.west) node[midway, above=0.12cm, font=\sffamily\small\bfseries, align=center] (lyrics-mlp) {Linear};
\node[midway, right=8.25cm of song] (fire1) {\includegraphics[height=0.45cm]{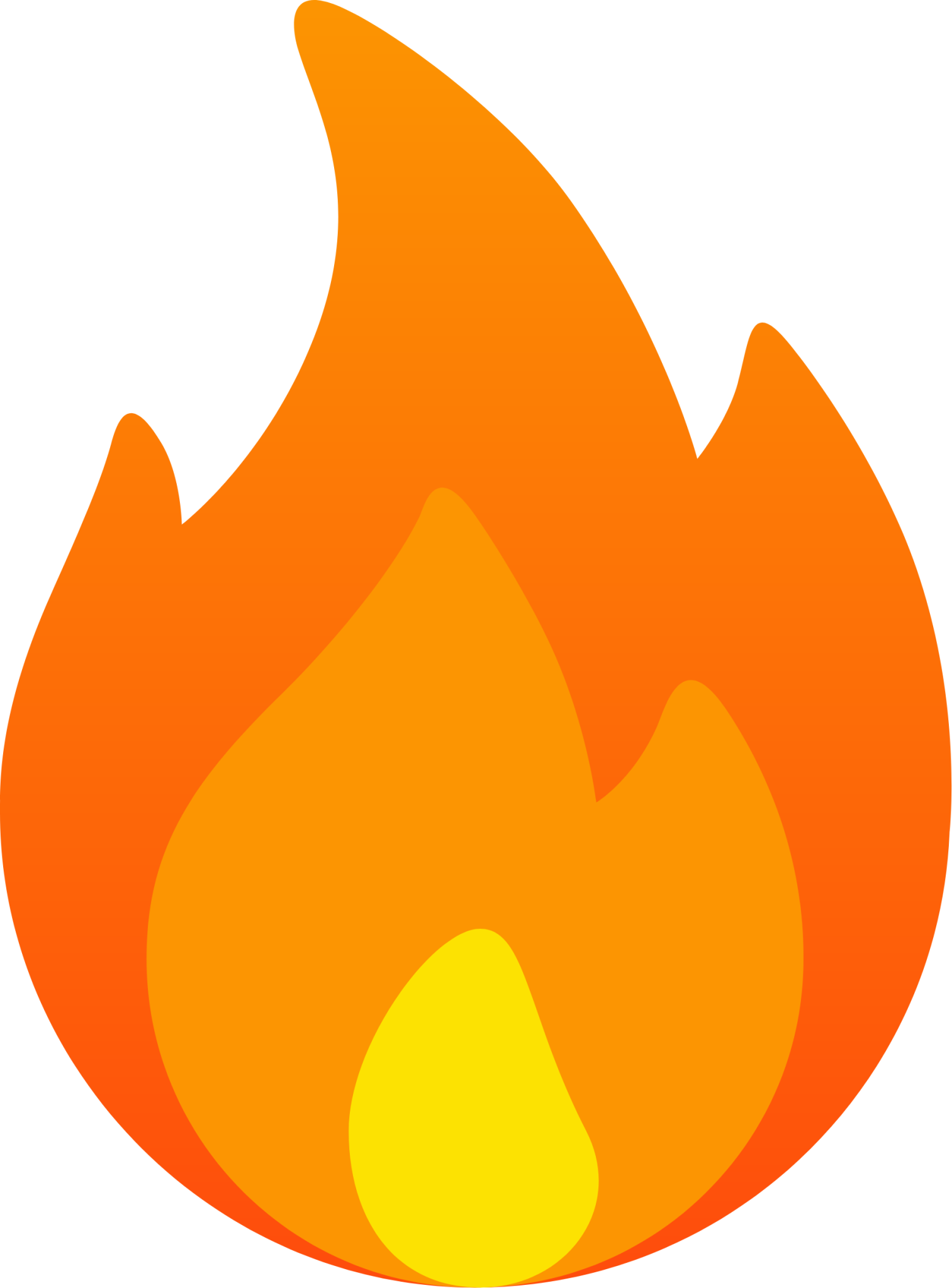}};

\node[feature, minimum width=0.8cm, minimum height=1.1cm, right=1.0cm of $(inter-top.east)!0.5!(inter-bottom.east)$] (mlp-block) {
  \begin{tabular}{@{} c @{}}
    $\cdot\cdot\cdot$ \\
    0.05 \\
    $\cdot\cdot\cdot$ \\
    0.45 \\
    $\cdot\cdot\cdot$
  \end{tabular}
};

\draw[line width=0.7pt] (inter-bottom.east) -- (mlp-block.north west);
\draw[line width=0.7pt] (inter-top.east) -- (mlp-block.south west);

\node[font=\sffamily\small, align=center, xshift=0.45cm] at ($ (inter-top.east)!0.5!(inter-bottom.east) $) (concat-label) {Concat};

\node[font=\sffamily\bfseries\normalsize, right=0.1cm of mlp-block, xshift=0.5cm] (mlp-label) {MLP};
\node[midway, right=11.6cm of song] (fire2) {\includegraphics[height=0.45cm]{figures/images/fire.png}};
\draw[arrow] (mlp-block.north east) -- ++(1, 0.35) node[decision, above right, fill=DeezerAcid] (real) {\textbf{Real}};
\draw[arrow] (mlp-block.south east) -- ++(1, -0.35) node[decision, below right, fill=DeezerOrange] (fake) {\textbf{Fake}};

\end{tikzpicture}
\caption{Overview of our pipeline to robustly detect AI-generated lyrics when only audio is available. In the top branch, we transcribe audio to lyrics via a \textit{transcriber}. This transcript is then used to get \textit{text features}. In the bottom branch, we use a speech model to get lyrics-related information only present in audio -- \textit{speech features}. Finally, we \textit{linearly project} and \textit{concatenate} both and feed them into an \textit{MLP} detector to classify the input song as \textcolor{green}{real} or \textcolor{red}{fake}. }
\label{fig:pipeline}
\end{figure*}

%% file: 2-method.tex
Existing unimodal approaches -- whether audio-based detectors that are sensitive to perturbations and generalize poorly, or lyrics-based detectors that require clean, often inaccessible lyrics -- tend to falter in real-world scenarios.
To address the impracticality of relying on perfectly clean lyrics, we turn to automatically transcribed lyrics.
However, transcripts capture \textit{what} (the semantic content), but they may miss  \textit{how} (subtle audio cues indicative of AI generation).
We hypothesize speech embeddings capture this \textit{how} -- lyrics-related cues present in audio but not in lyrics themselves.

To combine \textit{what} and \textit{how}, our method employs late fusion and synergistically integrates features from transcribed lyrics (semantic content) and speech (lyrics-related audio cues).
This multi-view fusion aims to overcome the limitations of text-only methods, enabling accurate detection resilient to audio attacks, as detailed in Section~\ref{sec:results}.
We provide an overview of our method in Figure~\ref{fig:pipeline}.

\vspace{1mm}
\noindent\textbf{(i) Text Branch.}
We use a transcription model (ASR model) to transcribe audio to lyrics.
To represent the semantic content of these lyrics for downstream processing, we feed the entire lyrics transcript into a text embedding model. This model captures rich semantic information and generates a single, contextualized \textit{lyrics text embedding} (top branch in Fig.~\ref{fig:pipeline}).

\vspace{1mm}
\noindent\textbf{(ii) Speech Branch.}
To capture the \textit{how} of lyrics (complementary audio cues indicative of AI generation) we use a speech model.
Unlike general audio embeddings, or the ASR models used for transcription, speech embedding models are specifically designed to capture rich acoustic and paralinguistic information from speech signals, such as prosody, intonation, and speaker characteristics, which can be indicative of AI generation even if not present in the transcribed text. To our knowledge, this is the first application of dedicated speech embeddings for AI-generated lyrics detection in the music domain.
This model extracts lyrics-related audio features: phonetic and contextual patterns like prosody and intonation from audio, resulting in a \textit{lyrics speech embedding} (bottom branch in Fig.~\ref{fig:pipeline}).
We also conducted in-domain experiments with source separation to isolate instances of singing voice. However, this did not significantly improve performance, suggesting our method is already somewhat resilient to background music.

\vspace{1mm}
\noindent\textbf{(iii) Late Fusion.}
We employ late fusion to synergistically combine lyrics and speech features, derived from audio alone.
Its simple and modular design offers key benefits: independent component updates,  preservation of each component's strengths (e.g., multilinguality), and robustness to component changes (cf.~§\ref{sec:component-instantiation}).
In the face of the evolving AIGM landscape, we argue these characteristics are crucial for a practically applicable robust detection system.
For fusion, features from both branches are linearly down-projected to $128$, concatenated, and then classified using a lightweight MLP, trained with binary cross-entropy loss (details in Appendix \ref{appendix:complete-experimental-results}).
Overall, our modular late-fusion design enables a robust, generalizable, and practically evolvable detection method.

%% file: 3-setup.tex
\input{figures/wer_comparison-plot}

\subsection{Dataset}
We start from the lyrics dataset of \citet{labrak2024detectingsyntheticlyricsfewshot}, which provides 3,655 real and 3,535 AI-generated lyrics from three LLM generators.
Human lyrics spanning nine languages and the six most popular genres per language are used as seeds in the generation pipeline.

A key limitation of this dataset is that it provides lyrics only.
Therefore, to enable realistic audio-based experiments representative of current AIGM, we generate corresponding audio for the AI-generated lyrics using state-of-the-art Suno v3.5, conditioned on lyrics and genre.\footnote{While tools such as Suno can also generate lyrics, we use the provided lyrics to ensure control over the lyrical content.}
For songs with human-generated lyrics, we use their original audio.
This results in a dataset of 7,190 songs, balanced between fully real songs and Suno-generated songs with AI lyrics generated by multiple LLMs.
We follow the train/test split of \citet{labrak2024detectingsyntheticlyricsfewshot}.

Moreover, a key question is whether our model and its components detect AI-generated lyrics or just audio artifacts inherent in AI-generated audio (robustness to AI audio artifacts). To address this, we design a "partly-fake" experiment:
generating Suno audio for \textit{real} lyrics and evaluating performance compared to detecting fully fake songs. 
This mitigates the influence of audio artifacts that should be mostly similar for partly-fake and fully fake, allowing us to verify if each model relies on lyrics.

In addition, to test generalization, we generate 260 additional songs with synthetic lyrics from the test set of the lyrics dataset by \citet{labrak2024detectingsyntheticlyricsfewshot} using \textit{Udio}, another music generation tool,\footnote{We searched for other models or tools capable of conditioning on lyrics but found no other suitable ones.} and sample 260 real songs.
Our model, trained on the Suno dataset, is then evaluated on these out-of-domain scenarios without further training.
For details, we refer to Appendix~\ref{appendix:complete-experimental-results}.

\input{figures/transcribers-plot}

\subsection{Evaluation Metrics}
Following \citep{labrak2024detectingsyntheticlyricsfewshot, nakov-etal-2013-semeval, li-etal-2024-mage}, we evaluate performance primarily via macro-recall as our main metric and additionally report AUROC scores.
We provide a definition of these metrics in Appendix \ref{appendix:metric-def}.

%% file: figures/wer_comparison-plot.tex
\begin{figure}[t]
    \center
    \centering
    \includegraphics[width=\linewidth]{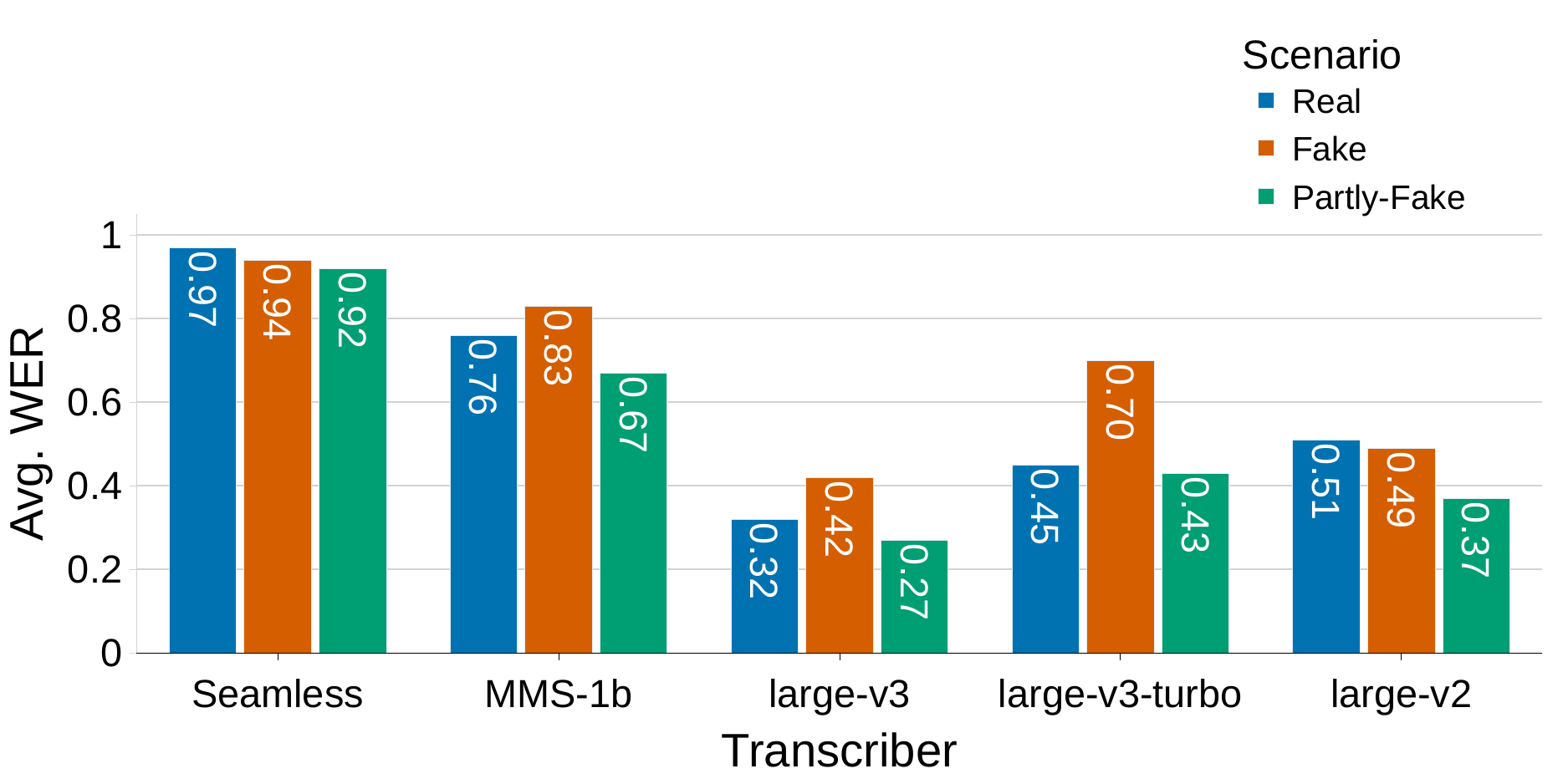}
    \caption{
        Word error rates (WER) of different transcription models across Real, Fake, and Partly-Fake song scenarios. Lower WER indicates better transcription.
    }
    \label{fig:transcibers-wer-comparison}
\end{figure}

%% file: figures/transcribers-plot.tex
\begin{figure}[t]
    \center
    \centering
    \includegraphics[width=\linewidth]{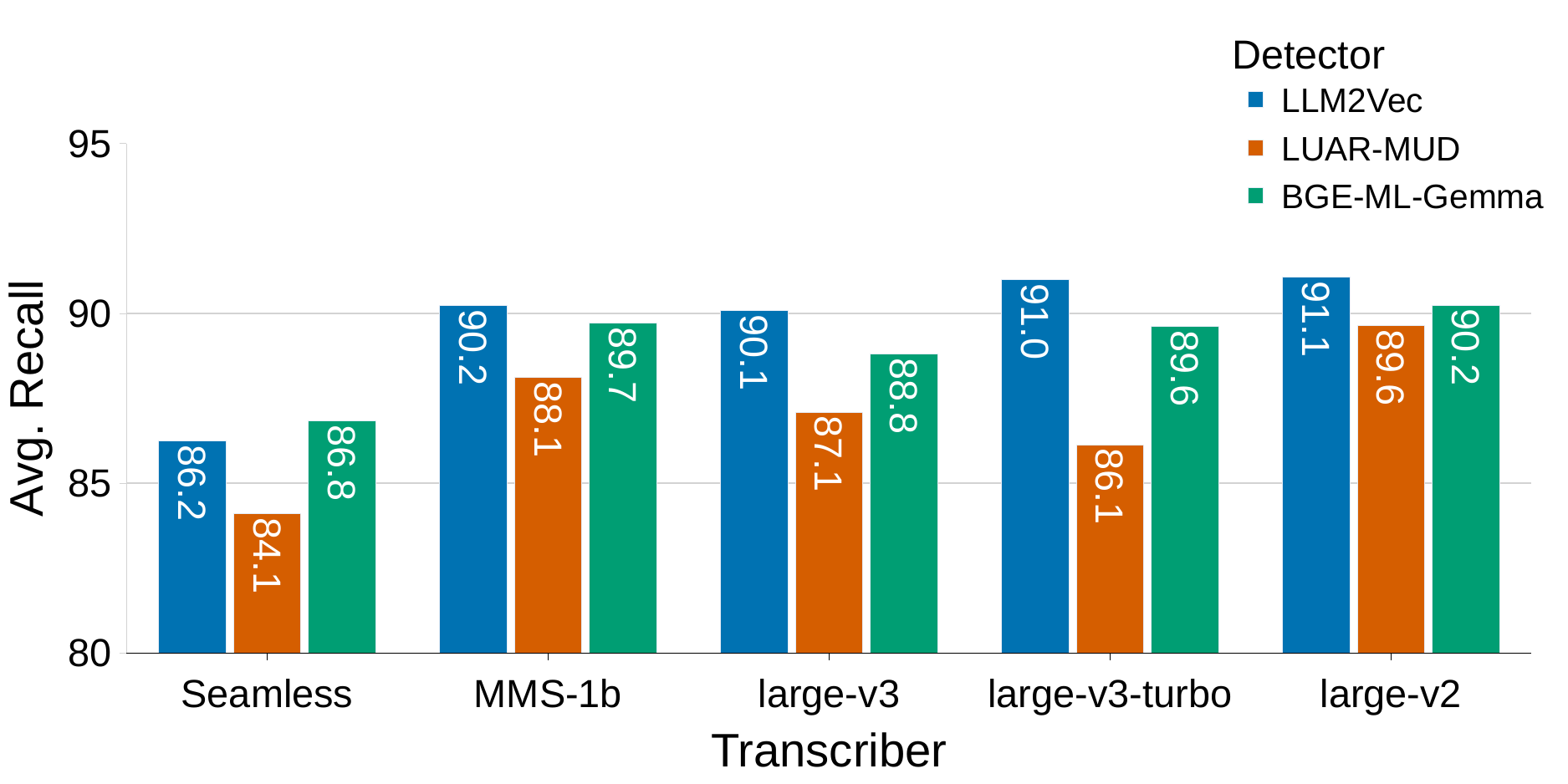}
    \caption{
        Recall of transcriber-feature combinations.
    }
    \label{fig:transcibers-comparison}
\end{figure}

%% file: 4-results.tex
\subsection{Component Instantiation}
\label{sec:component-instantiation}

We first evaluate several unimodal features to select as components in our multimodal pipeline, starting with text-based detectors.
To provide text for text-based detectors, we transcribe real and synthetic audio to lyrics using five recent multilingual transcription models:
Whisper in variations \textit{large-v2}, \textit{large-v3}, and \textit{large-v3-turbo}~\citep{Radford2022RobustSR}, \textit{mms-1b}~\citep{Pratap2023ScalingST}, and \textit{Seamless-large}~\citep{Communication2023SeamlessME}.

\paragraph{Transcription Quality (WER).} 
To first assess the intrinsic quality of these transcribers, we calculate their Word Error Rate (WER) against ground truth lyrics for different song types: human-generated (\textit{real}), AI-generated (\textit{fake}), and AI-generated audio with human lyrics (\textit{partly-fake}). Figure~\ref{fig:transcibers-wer-comparison} illustrates these WERs.
Overall, Whisper-based models demonstrate the highest transcription quality, with \textit{Whisper-large-v3} generally achieving the lowest WERs among all tested transcribers.
In contrast, \textit{MMS-1b} and \textit{Seamless} exhibit substantially higher WERs across all scenarios, indicating more transcription errors.
A notable pattern across Whisper models is that WERs on \textit{partly-fake} songs are often among the lowest, potentially due to the AI-generated audio in these cases being clearer or more consistently enunciated than some human recordings. Moreover, we also observe that \textit{real} songs were generally transcribed slightly worse than \textit{fake} songs across Whisper models.

\input{tables/component-init-table}

\paragraph{Impact on Downstream Detection.}
While the raw WER provides insights into transcriber quality, the key consideration is how these transcriptions affect performance on our downstream task of AI-generated lyrics detection. Therefore, we next evaluate the impact of using different transcribers when combined with various text-based detection features.
For text, we use the two best-performing detection features from \citet{labrak2024detectingsyntheticlyricsfewshot}: \textit{UAR-MUD}~\citep{rivera-soto-etal-2021-learning}, and \textit{LLM2Vec} with Llama3 8B as base model~\citep{behnamghader2024llm2veclargelanguagemodels}, and a recent multilingual general-purpose embeddings model, \textit{BGE-Multilingual-Gemma2}~\citep{chen-etal-2024-m3}.

Figure~\ref{fig:transcibers-comparison} compares the average detection performance of these transcriber-feature combinations, where we train an MLP classifier on each.
Results reveal that while Whisper-large-v3 exhibits slightly lower WERs (Figure~\ref{fig:transcibers-wer-comparison}), Whisper-large-v2 achieves the best average recall at $90.2$\%. This shows that lower raw WER does not necessarily correspond to improved detection performance.
Further, it also indicates that our feature extraction and classification pipeline can effectively handle the moderate level of transcription errors from models such as Whisper large-v2.
Similarly, text detector choice shows minimal difference: UAR-MUD performs slightly lower, while LLM2Vec shows the highest average recall.
Thus, robust detection performance is not tied to a single transcriber or text feature, indicating robustness of our approach to variations in unimodal components, even with the observed differences in raw WER.

Next, we evaluate speech embeddings from three strong multilingual models: \textit{XEUS}~\citep{chen-etal-2024-towards-robust}, \textit{Wav2Vec 2.0}~\citep{Baevski2020wav2vec2A}, and \textit{MMS-1b} (using the ASR-finetuned variant)~\citep{Pratap2023ScalingST}. For each, we apply mean-pooling to obtain a single vector.
As with text features, we train an MLP using the features.

Table~\ref{tab:component-init-table} shows results for each, and for comparison, includes text-based detector results with Whisper large-v2.
Comparing speech embeddings, performance margins are slightly larger, with \textsc{XEUS} performing best at $92.2$\% average recall.
We attribute its performance to a large and diverse training dataset that includes not only spoken dialogue but also instances of singing voice. This may enable the model to capture richer vocal characteristics relevant to distinguishing AI-generated sung lyrics, such as prosody and timbre. 
However, its training data lacks AI-generated voice, crucial for fair evaluation.
Given these findings, we use features from \textsc{LLM2Vec} with transcripts from Whisper large-v2 and \textsc{XEUS} to instantiate our multimodal pipeline.
In addition, Appendix~\ref{appendix:results-extra-features} shows results using various other text-based features.

\input{tables/artifacts-check-results}

\paragraph{Sensitivity to Audio Artifacts.}
We further analyze artifact influence using \textit{partly-fake} songs: Suno-generated audio with real lyrics.
Table~\ref{tab:artifacts-check-results} shows results for two scenarios: \textit{real vs. partly-fake} (differentiating human-generated vs. synthetic audio, both with human-generated lyrics) and \textit{fake vs. partly-fake} (differentiating synthetic vs. human-generated lyrics, both with AI-generated audio).

In the \textit{real vs. partly-fake} scenario,
the speech-based \textsc{XEUS} performs at a level consistent with random chance, indicating its features are not primarily driven by AI audio artifacts.
Transcription-based detectors, however, perform above random. This may be due to the transcription process capturing subtle audio generation artifacts (e.g., ASR training bias or differing distributions of non-lyrical tokens like ``[Outro]'').
Nevertheless, in the \textit{fake} vs. \textit{partly-fake} scenario (both audio types AI-generated, lyrics differ), performance is higher for all methods, with \textsc{XEUS} achieving $92.0$\% recall.
This suggests models primarily distinguish lyrical content even when audio is AI-generated, highlighting the resilience of our multi-view approach.

\input{tables/main-results-v2}

\subsection{In-domain Evaluation}
\label{sec:main-results}

Table~\ref{tab:overall-results} shows our main evaluation results on detecting AI-generated songs.
We compare our multi-view model (\textsc{XEUS}+\textsc{LLM2Vec} late fusion), which we term \textit{Double Entendre detect} (\textsc{DE-detect}), against the best unimodal baselines, and two additional strong baselines:
LLM2Vec (with Llama3 8B) using ground truth, non-transcribed lyrics, which was reported with high performance by \citet{labrak2024detectingsyntheticlyricsfewshot}, and a CNN trained on amplitude spectrograms to detect audio artifacts, following \citet{Afchar2024DetectingMD}.\footnote{Such models could also be trained on other input representations, but the findings of \citet{Afchar2024DetectingMD} are consistent across them, so we resort to the best-performing one.}

We first observe that $\textsc{Llama3 8B}_{\text{LLM2Vec}}$ using transcripts performs closely to $\textsc{GT Lyrics}_{\text{LLM2Vec}}$ (using clean, non-transcribed lyrics), reaching recall scores of $90.7$\% and $94.3$\%, respectively.
This indicates transcription effectively retains AI-generated lyric characteristics for detection.
Moreover, our multi-view model achieves higher scores than methods using audio-derived lyrics, reaching a recall of $94.9$\% (and an AUROC score of $98.5\%$), and even improves upon the clean ground truth lyrics baseline despite audio-only input. 
Only \textsc{CNN} slightly outperforms our method in-domain.

\input{tables/attack-results}

\subsection{Out-of-domain Evaluation}
We now evaluate robustness to (i) audio perturbations/attacks and (ii) out-of-domain generalization to Udio. 
The former simulates real-world audio variations and potential adversarial attacks, while the latter tests generalization w.r.t. audio generators.
Results are shown in Table~\ref{tab:attack-results}, painting a contrasting picture to in-domain findings:
The CNN shows large performance drops in attacks, especially pitch, and poor generalization to Udio ($56.9$\% recall), revealing its artifact sensitivity.
In contrast, models relying on lyrics-related information are much more stable, showing they are less prone to artifacts.
Finally, our multi-view model, \textsc{DE-detect}, shows recall scores $1.5$-$2$\% higher than the unimodal ones across these settings, suggesting consistently more robust performance, crucial for practical, real-world applications.
We also ablate different fusion components in Appendix \ref{appendix:mm-componentes-ablation}.

%% file: tables/component-init-table.tex
\begin{table}[t]{
\small
\centering
\setlength\tabcolsep{11pt} %
\setlength\extrarowheight{-2pt} %
\begin{tabular}{lc}

\specialrule{\lightrulewidth}{0em}{0em} %
\specialrule{\lightrulewidth}{0em}{0.5em} %

\textbf{Model} & \textbf{Recall} \\

\toprule

\multicolumn{2}{c}{Text-based Detectors via Whisper large-v2 Transcripts} \\
\midrule

\textsc{UAR-MUD} & 89.6 \\

\textsc{BGE-ML-Gemma} & 90.2 \\
$\textsc{Llama3 8B}_{\text{LLM2Vec}}$ & \textbf{90.7} \\

\midrule

\multicolumn{2}{c}{Speech-based Detectors} \\

\midrule

\textsc{Wav2Vec 2.0}  & 83.1 \\
\textsc{MMS-1b} & 88.8 \\
\textsc{XEUS} & \textbf{92.2} \\

\bottomrule

\end{tabular}%
}
\caption{Recall scores, macro-averaged over multilingual lyrics, for several unimodal detectors.}
\label{tab:component-init-table}
\end{table}

%% file: tables/artifacts-check-results.tex
\begin{table}[t]{
\small
\centering
\setlength\extrarowheight{-2pt}
\setlength\tabcolsep{7pt}
\centering
\begin{tabular}{lccc}

\specialrule{\lightrulewidth}{0em}{0em} %
\specialrule{\lightrulewidth}{0em}{-0.2em} %

\textbf{Model} & \thead{\textsc{Real vs.} \\ \textsc{Partly-Fake}} & \thead{\textsc{Fake vs.} \\ \textsc{Partly-Fake}} \\
\toprule
\textsc{UAR-MUD} & \textbf{66.9} & 86.1 \\
\textsc{Llama3 8B} & 64.9 & 90.0 \\
\textsc{BGE-Ml-Gemma} & 67.7 & 89.0 \\

\midrule

\textsc{Wav2Vec 2.0} & \textbf{50.9} & 83.1 \\
\textsc{MMS-1b} & 50.7 & 88.5 \\
\textsc{XEUS} & 50.5 & \textbf{92.0} \\

\bottomrule

\end{tabular}%
}
\caption{Recall scores on detecting \textit{partly-fake} songs with human-generated lyrics but synthetic audio.  
}
\label{tab:artifacts-check-results}
\end{table}

%% file: tables/main-results-v2.tex
\begin{table}[t]{
\small
\centering
\setlength\tabcolsep{8.9pt}
\setlength\extrarowheight{-2pt}
\centering
\begin{tabular}{lcccc}

\specialrule{\lightrulewidth}{0em}{0em} %
\specialrule{\lightrulewidth}{0em}{0.5em} %
\textbf{Model} & \multicolumn{2}{c}{\textbf{Recall}} &  \multicolumn{2}{c}{\textbf{AUROC}} \\
&  \texttt{en} & \texttt{all} &  \texttt{en} & \texttt{all} \\

\midrule

$\textsc{GT Lyrics}_{\text{LLM2Vec}}${\textbf{\textdagger}}
 & 91.3 & 94.3 & 99.0 & 97.3\\
 
$\textsc{CNN}_{\text{Spectrogram}}${\textbf{\textdaggerdbl}} & \textbf{97.5} & \textbf{97.4} & \textbf{99.9} & \textbf{99.8} \\

\cdashline{1-5}
\vspace{-2mm} \\

\textsc{XEUS} & 89.1 & 92.2 & 94.5 & 97.0 \\

$\textsc{Llama3 8B}_{\text{LLM2Vec}}$ & 90.6 & 90.7 & 97.6 & 94.8\\

\textsc{\textbf{DE-detect}} & \textbf{93.9} & \textbf{94.9} & \textbf{98.2} & \textbf{98.5} \\
\bottomrule

\end{tabular}%
}
\caption{Recall and AUROC scores on English-language and macro-averaged over multilingual lyrics.
For transcription, we use Whisper large-v2. For \textsc{Ours}, we combine embeddings from LLM2Vec and XEUS.
\textsuperscript{\textbf{\textdagger}}~denotes the best-performing baseline by \citet{labrak2024detectingsyntheticlyricsfewshot}, using non-transcribed ground truth (GT) lyrics with $\textsc{Llama3 8B}_{\text{LLM2Vec}}$.
\textsuperscript{\textbf{\textdaggerdbl}}~uses the amplitude spectrogram to train a CNN on the task as in \citep{Afchar2024DetectingMD}.
}
\label{tab:overall-results}
\end{table}

%% file: tables/attack-results.tex
\begin{table}[!t]{
\small
\centering
\setlength\extrarowheight{-2pt}
\setlength\tabcolsep{4.1pt}
\centering
\begin{tabular}{lccccc@{\hspace{1.5mm}\hspace{1.5mm}}c} %

\specialrule{\lightrulewidth}{0em}{0em} %
\specialrule{\lightrulewidth}{0em}{0.5em} %
& \multicolumn{5}{c}{\textsc{Audio Attacks}} &  \\ 
\cmidrule(lr){2-6}
\multirow{2}{*}[1.5em]{\textbf{Model}} 
& \scalebox{0.83}{{Stretch}} & \scalebox{0.83}{{Pitch}} & \scalebox{0.83}{{EQ}} & \scalebox{0.83}{{Noise}} & \scalebox{0.83}{{Reverb}} 
& \multirow{2}{*}[1.5em]{\textsc{Udio}} \\
\midrule
\textsc{CNN} & \textbf{98.1} & 59.0 & 79.4 & 77.4 & 80.7 & 56.9 \\

\cdashline{1-7}
\vspace{-2mm} \\
\textsc{XEUS} & 92.5 & 92.3 & 92.3 & 92.4 & 92.4 & 85.9 \\
\textsc{UAR-MUD} & 86.7 & 88.8 & 88.8 & 88.6 & 88.5 & 85.6 \\
\textsc{Llama3 8B} & 90.0 & 89.7 & 89.6 & 89.3 & 89.6 & 85.9 \\

\textsc{\textbf{DE-detect}} & 94.1 & \textbf{93.9} & \textbf{94.0} & \textbf{93.9} & \textbf{94.1} & \textbf{87.9} \\

\bottomrule

\end{tabular}%
}
\caption{Recall scores on out-of-distribution data (Udio) and when fake songs are perturbed (attacked) in five different ways. We report average scores over languages.}
\label{tab:attack-results}
\end{table}

%% file: 9-conclusion.tex
In this work, we proposed a novel modular multimodal approach --  \textit{Double Entendre detect} (\textsc{DE-detect}) -- for robust AI-generated lyrics detection, late-fusing lyrics and speech representations.
\textsc{DE-detect} consistently outperformed text-based baselines in-domain.
We also stressed the importance of robustness for practical AIGM detection
and showed that our method is more robust than all unimodal ones.
Our findings underscore the importance of considering both lyrical and speech features for reliable detection,
offering a more resilient and forward-looking solution with significant implications for copyright, music industry transparency, and the evolving relationship between humans and AI in creative domains.

%% file: 999-appendix.tex
\section{Complete Experiment Details}
    \input{tables/training-overview}
    \vspace{2mm}
    \noindent\textbf{Training Overview.} For clarity, we provide an overview of the training pipeline for each model type in Table \ref{tab:training_overview}.
    
\label{appendix:complete-experimental-results}
    \vspace{2mm}
    \noindent\textbf{Computing Infrastructure.}
    We transcribe lyrics and compute their features on a server with a single Nvidia RTX A5000 GPU and Intel Xeon Gold 6244 CPUs. We also use it for training lightweight MLPs and the $\textsc{CNN}_{\text{Spectrogram}}$ baseline.

    \vspace{2mm}
    \noindent\textbf{Implementation Details.}
    We use the \texttt{PyTorch}~\citep{NEURIPS2019_9015} and \texttt{transformers}~\citep{wolf-etal-2020-transformers} libraries and use models in fp16 for all experiments.
    To make sure no encoding-specific patterns are picked up, we convert all audio to mp3 with 128kbps.

    \vspace{2mm}
    \noindent\textbf{Transcription.}
    To transcribe audio to lyrics via Whisper models, we use the \texttt{faster-whisper} \citep{guillaume_2023_fasterwhisper}.
    For \textit{mms-1b} and \textit{Seamless}, we use the transformers library with model versions \texttt{facebook/mms-1b-all} and \texttt{facebook/hf-seamless-m4t-large}, respectively. Since both models require language codes and utilize language adapters, we use the language identification module of Whisper large-v3 to provide the required language code. We also transcribed when using the ground truth language code (which, however, is unrealistic in practical scenarios), but did not find it to consistently improve transcription performance.
    Additionally, we experimented with applying source separation but did not find it improves performance, which is in line with the findings of \citet{Cfka2024LyricsTF}.

    \vspace{2mm}
    \noindent\textbf{Audio-based Baselines.}
    We first convert the waveform to mono in 16kHz, the input format of our speech embedding models, to extract speech features.
    To use \textsc{XEUS}, we use the ESPnet library \citep{Watanabe2018ESPnetES} and disable masking.
    For \textsc{Wav2vec 2.0} and \textsc{MMS-1b}, we use the \texttt{transformers}~\citep{wolf-etal-2020-transformers} library with model versions \texttt{facebook/wav2vec2-large-960} and \texttt{facebook/mms-1b-all}, respectively. 
    Since all models extract several feature vectors whose size depends on the duration of the audio sample, we apply mean-pooling to aggregate these features into a single, fixed-length \textit{speech embedding}.
    We also experimented with source separation but observed that it resulted in similar detection performance with worse generalization. This indicates that, indeed, semantics of sung lyrics are being captured, and that source separation is not robust w.r.t. audio artifacts.
    
    \vspace{2mm}
    \noindent\textbf{Text-based Baselines.}
    For \textsc{LLM2Vec}, we use \texttt{McGill-NLP/LLM2Vec-Meta-Llama-3-8B}-\texttt{Instruct-mntp}, i.e., the mntp-tuned (masked next token prediction) of Llama3 8B, following \citet{labrak2024detectingsyntheticlyricsfewshot}.
    For \textsc{MiniLMv2}, \textsc{BGE-M3}, and \textsc{BGE-ML-Gemma}, we utilize the \texttt{sentence-transformers} \citep{reimers-gurevych-2019-sentence} library with model versions \texttt{sentence-transformers/all-MiniLM-L6-v2}, \texttt{BAAI/bge-m3}, and \texttt{BAAI-bge-multilingual}- \texttt{gemma2}, respectively.
    Finally, for UAR models, we use \texttt{UAR-MUD} and \texttt{UAR-CRUD}, respectively.
    To stay within memory constraints, we truncate the input to each model to a maximum of $512$ tokens. Note that this only affects a handful of songs.

    \vspace{2mm}
    \noindent\textbf{Audio generation.}
    To generate songs with Suno, we use their latest stable audio generation model, v3.5.
    Crucially, unlike previous versions that can only generate relatively short songs, it can create songs with up to 4 minutes, making them much more realistic.
    Specifically, we copy the LLM-generated lyrics into the \textit{Lyrics} field and the song's corresponding genre into the \textit{Style of Music} field. We follow this process using both synthetic and human-written lyrics. For the latter, a few songs were blocked during generation, making our \textit{Partly-Fake} subset slightly smaller than the human-written one.

    For our Udio subset used to test generalization, we use the latest and highest-quality udio-130 v1.5 model. We copy the LLM-generated lyrics for the stratified subset of 260 samples from the test set of lyrics into the \textit{Lyrics Editor} field and fill the song's genre to \textit{Describe your Song}. For controllability, we set \textit{Lyrics Strength} to 100\% and the seed to $42$, leaving the rest unchanged. Since Udio does not support generating songs with real lyrics (i.e., \textit{Partly-Fake}), we could not consider this scenario.

    Since both Suno and Udio generate two songs with different audio per generation requests, we compute features, train models, and evaluate both independently, and then average over them.

    \vspace{2mm}
    \noindent\textbf{Audio perturbations.}
    We use pedalboard \citep{sobot_peter_2023_7817838} and librosa \citep{McFee2015librosaAA} to perturb audio. To simulate real-life audio attacks, we only perturb AI-generated audio and base our implementation on \citet{Afchar2024DetectingMD}.

    \vspace{2mm}
    \noindent\textbf{MLP training.}
    To evaluate unimodal features, we train a multi-layer perceptron (MLP) with two hidden layers of size 256 and 128, respectively, and ReLU activation function. 
    For the multimodal fusion MLP, we first project each feature to an intermediate representation of size 128. After concatenation, we apply a ReLU activation function and a linear layer with size 128 before the final classification layer.
    They are each optimized with AdamW \citep{Loshchilov2017DecoupledWD} optimizer with a learning rate of 1e-3, scaled down by a factor of $0.1$ if the training loss does not increase for five consecutive epochs.
    We also experimented with different settings and classifiers, such as kNN, but noticed that the specific configuration of both unimodal and multimodal MLPs does not make a significant difference in detection results. For MLP training, we use \texttt{pytorch-lightning} \citep{Falcon_PyTorch_Lightning_2019}, a wrapper of \texttt{PyTorch} \citep{NEURIPS2019_9015}.

\section{Information about Feature Extractors}
    We distinguish between models employed for \textit{Automatic Speech Recognition (ASR)} to obtain lyric transcripts (for the text branch) and models used to extract \textit{speech embeddings} that capture acoustic and paralinguistic cues directly from the audio (for the speech branch).

    \noindent\textbf{Text Features.}
    \textsc{LLM2Vec}~\citep{behnamghader2024llm2veclargelanguagemodels} is an unsupervised method transforming autoregressive LLMs into text encoders in a three-step process. 
    First, bidirectional attention is enabled by modifying the causal attention mask to a bidirectional one.
    The next step is masked next-token prediction (MNTP), where the model is trained on a small dataset to adapt it to this new attention mask.
    The final, optional step consists of SimCSE~\citep{gao-etal-2021-simcse} learning, where the model is adapted on larger, more diverse datasets to improve sequence representation for downstream tasks.

    Universal Authorship Attribution models~\citep[\textsc{UAR};][]{rivera-soto-etal-2021-learning} capture capture authorial writing style.
    They exist in variants MUD (\textsc{UAR-MUD}) and CRUD (\textsc{UAR-CRUD}), trained on texts from 1 million and 5 Reddit users, respectively.

    Finally, \textsc{BGE-ML-Gemma} adapts Gemma2 9B \citep{Riviere2024Gemma2I} to a multilingual text embedded using the M3-Embedding methodology by \citet{chen-etal-2024-m3} on diverse multilingual datasets, resulting in a strong general text embedding model, particularly excelling in multilingual tasks.

    \vspace{1mm}
    \noindent\textbf{Speech Features.}
    \textsc{Wav2Vec 2.0}~\citep{Baevski2020wav2vec2A} uses self-supervised learning to learn speech representations from raw audio.
    It uses a convolutional network to create latent representations and a Transformer to build contextualized representations.
    Pre-training involves identifying masked quantized latent representations, enabling powerful representations from unlabeled data for downstream speech tasks.

    Next, \textsc{mms-1B}~\citep{Pratap2023ScalingST} is a multilingual speech model that supports speech in over 1,000 languages.
    It expands the number of supported languages by over 40x, trained using self-supervised learning with Wav2vec 2.0 using data unlabeled from publicly available religious texts.
    For its use in our \textit{speech branch} (i.e., for extracting embeddings), we utilize an \textit{ASR-finetuned variant} of \textsc{mms-1b}. We leverage the encoder outputs from this variant to capture rich acoustic and paralinguistic features relevant to sung speech, rather than its final transcribed text output. This application is distinct from using \textsc{mms-1b} as a full ASR system for transcription, a role in which we also evaluate it (c.f. Section~\ref{sec:component-instantiation}).

    Finally, \textsc{XEUS}~\citep{chen-etal-2024-towards-robust} represents the current state-of-the-art in multilingual speech representation learning, extending language coverage four-fold by combining speech from publicly accessible corpora with a newly created corpus of 7400+ hours from 4,057 languages.
    Moreover, a novel joint dereverberation task is introduced to improve robustness.

\section{Ablation Study}
\label{appendix:mm-componentes-ablation}

\input{tables/ablate-mm-components-v2}
    In Table~\ref{tab:ablate-mm-components-results}, we ablate the choice of fusing speech and transcript-based lyrics embeddings from \textsc{XEUS} and $\textsc{Llama3 8B}_{\text{LLM2Vec}}$, respectively.
    We late-fuse two of each of the best-performing text and speech features both in unimodal and multimodal combinations, resulting in our model, \textsc{DE-detect}
    While some unimodal combinations improve performance compared to only using one feature, none reaches our multimodal model's performance.
    Moreover, other multimodal feature combinations get close to the performance of \textsc{DE-detect} (e.g., \textsc{XEUS+UAR-MUD}), none outperforms \textsc{DE-detect}. However, multimodal methods consistently outperform their unimodal counterparts.
    Overall, this further demonstrates the robustness of our pipeline with respect to different components.

\section{Results using Additional Features}
\label{appendix:results-extra-features}
   \input{tables/llm-prob-extra-results-v2}
    Furthermore, we show results using additional neural features and several probabilistic features based on Llama3 8B per-tokens probabilities in Table \ref{tab:llm-prob-extra-results}.
    For neural features, we use another variation of UAR, \textsc{UAR-CRUD}, trained on a smaller dataset \citep{rivera-soto-etal-2021-learning}. Moreover, we evaluate two more text embedding models, \textsc{MiniLM-L6-v2}~\citep{wang-etal-2021-minilmv2}, an efficient lightweight model, as well as another recent strong text embedders, \textsc{BGE-M3}~\citep{chen-etal-2024-m3}.
    \textsc{Perplexity} \citep{10.1007/978-3-319-41754-7_43} corresponds to the overall likelihood of the lyrics based on an exponential average using the negative log-likelihood (NLL).
    Shannon \textsc{Entropy} \citep{Shannon1948math, 10.5555/3053718.3053722} measures the diversity of text leveraging token-level NLL.
    \textsc{Min-K\% Prob} \citep{shi2024detecting} selects a subset of the lowest token-level NLL values, with the size of the subset being K\%. We use $K=10$, following \citet{labrak2024detectingsyntheticlyricsfewshot}.
    Finally, \textsc{Max. Neg. LL} \citep{10.5555/3618408.3619446, solaiman2019releasestrategiessocialimpacts, gehrmann-etal-2019-gltr, ippolito-etal-2020-automatic} uses the maximum token-level NLL as a single feature.

\section{Effect of Different Transcribers}
\label{appendix:effect-transcribers}
    We show complete results on a non-Whisper transcriber, \textit{MMS-1b}, in Table~\ref{tab:overall-results-meta-mms}, demonstrating similar performance and patterns as with using \textit{Whisper large-v2}.
    This further demonstrates our method is not reliant on any specific architecture for its transcription component.

   \input{tables/main-results-meta_mms-v2}

\section{Metrics definition}
\label{appendix:metric-def}

\paragraph{Macro-Recall.}
Given a binary classification task with classes $C = \{c_1, c_2\}$ (in our case, real and AI-generated), recall for a specific class $c_i$ is defined as:
$$ \text{Recall}(c_i) = \frac{\text{TP}_i}{\text{TP}_i + \text{FN}_i} $$
where $\text{TP}_i$ is the number of true positives for class $c_i$ (samples correctly identified as $c_i$), and $\text{FN}_i$ is the number of false negatives for class $c_i$ (samples of $c_i$ incorrectly identified as belonging to another class).
Macro-recall is then the unweighted arithmetic mean of the per-class recalls:
$$ \text{Macro-recall} = \frac{1}{|C|} \sum_{i=1}^{|C|} \text{Recall}(c_i) $$
This metric is chosen as it gives equal weight to the performance on each class, which is crucial for tasks where misclassification costs might be similar for all classes or when class imbalance is present, ensuring that the performance on a minority class is not overshadowed.

\paragraph{AUROC.}
The AUROC quantifies the overall ability of a classifier to discriminate between positive and negative classes across various decision thresholds. It is the area under the ROC curve, which plots the true positive rate (TPR, equivalent to recall or sensitivity) against the false positive rate (FPR) at different threshold settings.
$$ \text{TPR} = \frac{\text{TP}}{\text{TP} + \text{FN}} $$
$$ \text{FPR} = \frac{\text{FP}}{\text{FP} + \text{TN}} $$
where TP, FN, FP (False Positives), and TN (True Negatives) are defined with respect to a designated positive class (e.g., AI-generated). An AUROC of 1.0 signifies a perfect classifier, correctly distinguishing all positive and negative instances, while an AUROC of 0.5 suggests performance no better than random guessing.

%% file: tables/training-overview.tex
\begin{table*}[h]
\centering
\small
\begin{tabular}{@{}llll@{}}
\toprule
\textbf{Model Type}         & \textbf{Input} & \textbf{Processing Pipeline} & \textbf{Classifier}                                                                                                \\ 
\midrule
CNN Baseline         & Audio waveform          & Amplitude spectrogram $\Rightarrow$ CNN  & - \\
GT Lyrics Baseline   & Ground truth lyrics     & Text embedding model (LLM2Vec)         & MLP (256, ReLU, 128, 2)  \\     
\midrule
Unimodal (Text)   & Audio waveform      & Transcriber $\Rightarrow$ Text embedding model    & MLP (256, ReLU, 128, 2) \\
Unimodal (Speech) & Audio waveform          & Speech embedding model $\Rightarrow$ Mean pooling  & MLP (256, ReLU, 128, 2) \\
\midrule
\textsc{DE-detect} (multi-view)    & Audio waveform          & \makecell[l]{Transcriber (Whisper) $\Rightarrow$ Text Embedding \\ Speech embedding $\Rightarrow$ Mean Pooling \\ Linearly project both to 128, concatenation} & MLP (128, ReLU, 128, 2) \\
\bottomrule
\end{tabular}
\caption{Training overview for each model type. Each is trained on the same set of songs in each scenario. For models other than CNN, only the MLP classifier is trained, while the rest of the processing pipeline remains frozen.}
\label{tab:training_overview}
\end{table*}

%% file: tables/ablate-mm-components-v2.tex
\begin{table}[!t]{
\small
\centering
\begin{tabular}{lcc}

\textbf{Model} &  \texttt{en} & \texttt{all} \\

\toprule

\multicolumn{3}{c}{\textsc{Speech Embeddings}} \\
\midrule
\textsc{Wav2Vec 2.0}  & 78.2 & 83.1 \\

\textsc{MMS-1b} & 80.7 & 88.8 \\
\textsc{XEUS} & 89.1 & 92.2 \\

\cdashline{1-3}
\vspace{-2mm} \\

\textsc{Wav2vec 2.0 + MMS-1b} & 87.8 & 93.2 \\

\textsc{Wav2vec 2.0 + XEUS} & 87.7 & 92.2 \\

\textsc{XEUS + MMS-1b} & 87.8 & 93.2 \\

\midrule
\multicolumn{3}{c}{\textsc{Text-based Detectors (Lyrics Transcription)}} \\
\midrule

\textsc{UAR-MUD} & 85.2 & 89.6 \\
\textsc{BGE-ML-Gemma} & 84.4 & 90.2 \\
$\textsc{Llama3 8B}_{\text{LLM2Vec}}$ & 90.6 & 90.7 \\

\cdashline{1-3}
\vspace{-2mm} \\
\textsc{UAR-MUD + BGE-ML-Gemma} & 84.0 & 90.0 \\
\textsc{UAR-MUD + LLM2Vec} & 91.2 & 92.2 \\
\textsc{BGE-ML-Gemma + LLM2Vec} & 87.4 & 91.7 \\

\midrule
\multicolumn{3}{c}{\textsc{Multimodal}} \\
\midrule
\textsc{\textbf{XEUS+LLM2Vec} \textbf{(Ours)}} & \textbf{93.9} & \textbf{94.9} \\
\textsc{XEUS+UAR-MUD} & 92.0 & 94.3 \\

\textsc{XEUS+BGE-ML-Gemma} & 91.8 & 94.0 \\

\cdashline{1-3}

\textsc{Wav2Vec 2.0+LLM2Vec}  & 92.2 & 92.9 \\
\textsc{Wav2Vec 2.0+UAR-MUD} & 85.9  & 90.5 \\

\textsc{Wav2Vec 2.0+BGE-ML-Gemma} & 88.5 & 92.0 \\

\cdashline{1-3}
\vspace{-2mm} \\

\textsc{MMS-1b+LLM2Vec}  & 91.8 & 93.1 \\

\textsc{MMS-1b+UAR-MUD} & 88.1 & 91.1 \\

\textsc{MMS-1b+BGE-ML-Gemma} & 87.2 & 92.1 \\

\bottomrule

\end{tabular}%
}
\caption{Recall scores on English-language songs and macro-averaged over multilingual lyrics using different unimodal and multimodal feature combinations. For transcription, we use Whisper-large-v2.
}
\label{tab:ablate-mm-components-results}
\end{table}

%% file: tables/llm-prob-extra-results-v2.tex
\begin{table}[!t]{
\small
\centering
\begin{tabular}{lcc}

\textbf{Model} &  \texttt{en} & \texttt{all}  \\

\toprule

\multicolumn{3}{c}{\textsc{Text-based Detectors (Lyrics Transcription)}} \\
\midrule

 \multicolumn{3}{c}{\textit{Neural Embeddings}} \\

\textsc{UAR-CRUD} & 81.9  & 88.2 \\

\textsc{MiniLMv2}  & 80.8 & 87.3 \\

\textsc{BGE-M3}  & 84.7 & 87.7 \\

\textsc{BGE-ML-Gemma} & 84.4 & 90.2 \\

\cdashline{1-3}

 \vspace{-2mm}\\
 \multicolumn{3}{c}{\textit{Metrics based on Llama3 8B Per-Tokens Probabilities}} \\

\textsc{Perplexity} & 53.4 & 34.9 \\

\textsc{Max. Neg. LL} & 61.4 & 55.8 \\

\textsc{Shannon Entropy} & 56.5 & 59.8 \\

\textsc{Min-K\%Prob (K=10)} & 66.0 & 54.0 \\
\bottomrule

\end{tabular}%
}
\caption{Recall scores on English songs and macro-averaged over multilingual lyrics using additional neural and probabilistic features based on Llama3 8B per-tokens probabilities using Whisper large-v2 transcripts.
}
\label{tab:llm-prob-extra-results}
\end{table}

%% file: tables/main-results-meta_mms-v2.tex
\begin{table}[!t]{
\small
\centering
\begin{tabular}{lcc}

\textbf{Model} &  \texttt{en} & \texttt{all} \\

\toprule

\multicolumn{3}{c}{\textsc{Speech Embeddings}} \\
\midrule
\textsc{Wav2Vec 2.0}  & 78.2 & 83.1 \\

\textsc{MMS-1b} & 80.7 & 88.8 \\
\textsc{XEUS} & \textbf{89.1} & \textbf{92.2} \\

\midrule
\multicolumn{3}{c}{\textsc{Text-based Detectors (Lyrics Transcription)}} \\
\midrule

\textsc{UAR-CRUD} & 78.8 & 88.1 \\

\textsc{UAR-MUD} & 78.0 & 88.1 \\

\textsc{MiniLMv2} & 81.1 & 87.7 \\

\textsc{BGE-M3} & 81.9 & 87.7 \\

\textsc{BGE-ML-Gemma} & \textbf{85.4} & 89.7 \\

$\textsc{Llama3 8B}_{\text{LLM2Vec}}$ & \textbf{85.4} & \textbf{90.3} \\

\midrule
\multicolumn{3}{c}{\textsc{Multimodal}} \\
\midrule
\textsc{\textbf{DE-detect}} & \textbf{89.1} & \textbf{93.6} \\

\bottomrule

\end{tabular}%
}
\caption{Recall scores on English-language songs and macro-averaged over multilingual lyrics using a different transcriber, \textit{MMS-1b}.
In this setting, \textsc{DE-detect} combines XEUS embeddings with $\textsc{Llama3 8B}_{\text{LLM2Vec}}$ embeddings from \textit{MMS-1b} transcriptions. While speech embeddings' scores remain unchanged when changing the transcriber, we include them for completeness.
}
\label{tab:overall-results-meta-mms}
\end{table}